# Pan-Tilt Camera and PIR Sensor Fusion Based Moving Object Detection for Mobile Security Robots

YongChol Sin, MyongSong Choe, GyongIl Ryang

Faculty of Electronics & Automation, Kim Il Sung University, Pyongyang,

Democratic People's Republic of Korea

**Abstract:** One of fundamental issues for security robots is to detect and track people in the surroundings. The main problems of this task are real-time constraints, a changing background, varying illumination conditions and a non-rigid shape of the person to be tracked. In this paper, we propose a solution for tracking with a pan-tilt camera and a passive infrared range (PIR) sensor to detect the moving object based on consecutive frame difference. The proposed method is excellent in real-time performance because it requires only a little memory and computation. Experiment results show that this method can detect the moving object such as human efficiently and accurately in non-stationary and complex indoor environment.

**Key Words**: mobile robot tracking, pan-tilt camera, human detection

## I. INTRODUCTION

In recent years, there are many works of the so-called human centered robotics and human–robot interaction and it is one of the most interesting research fields in mobile robotics. In general, a service robot has to focus its attention on humans and be aware of their presence. It is necessary, therefore, to have a tracking system that returns the current position, with respect to the robot, of the adjacent persons. This is a very challenging task, as people's behaviors are often completely unpredictable. Researchers have been using different methods to deal with this problem, in many cases, with solutions subjected to strong limitations, such as tracking in rather simple situations with a static robot or using some additional distributed sensors in the environment.

The necessity of fast and reliable systems for tracking people with mobile robots is evidenced in the literature by the growing number of real-world applications. Human tracking can help security robots plan that adapt their movements according to the motion of the adjacent people or follow an instructor across different areas of a building.[15,16,17]

For example, the tour-guide robot of Burgard *et al.* [2] adopts laser-based people tracking both for interacting with users and for mapping the environment, discarding human occlusions. Another example is the system implemented by Liu *et al.* [3], where a mobile robot tracks possible intruders in a restricted area and signals their presence to the security personnel.

Early, much work in the field of person detection and tracking is based on vision systems using only stationary cameras [4]. During the past decades, researchers in vision technique have already proposed various algorithms for detecting moving objects, such as, consecutive temporal difference (consecutive frames subtraction) [5,6,7,14], optical flow approach [8,9], and background subtraction [10,11], etc. Among these methods, background subtraction algorithms are most popular, because they are relatively simple in computing in a static scene.

However, the background is assumed to be static in this method. Thus, shaking cameras, waving trees, lighting changes are quite probable to cause serious problems to a background subtraction model . In addition, a successful background subtraction method needs to model the background as accurate as possible, and to adapt quickly to the changes in the background. These requirements add extra complexity to the computation of the model and make a real-time detection difficult to achieve.

Optical flow approach is quite excellent because it can detect the moving objects independently and it works very well in changing environments, even in the absence of any previous information of the background. However, the computational cost of the approach is very expensive, which makes it very difficult to be applied in a real-time system. Temporal difference is the simplest method to extract moving objects and robust to dynamic environments. However, it easy to cause small holes and cannot detect the entire shape of a moving object with uniform intensity. Also, any changing elements in the background can be easily classified as the foreground by temporal difference.

Most of them originate from virtual reality applications where one person moves in front of a stationary background. Moving objects are detected by subtracting two frames. These approaches often require that the tracked person is never occluded by other objects. Therefore they are typically not suitable for person following on a mobile robot since at the same time the robot tracks the person, it has to move into the person's direction to stay close enough.[16,19]

The most suitable devices used for people tracking are laser sensors and cameras. For instance, Lindström and Eklundh [12] propose a laser-based approach to track a walking person with a mobile robot. The system detects only moving objects, keeping track of them with a heuristic algorithm, and needs the robot to be static or move very slowly. Zajdel *et al.* [13] illustrate a vision based tracking and identification system which makes use of a dynamic Bayesian network for handling multiple targets. Even in this case, targets can be detected only when moving. Moreover, the working range is limited by the camera's angle of view; hence, it is difficult to track more than two subjects at the same time.

The solution presented in this paper adopts multi-sensor data fusion techniques for tracking people from a mobile robot combining a pan-tilt camera and a passive infrared range sensor. A new detection algorithm has been implemented to find human body by using 3 PIR sensors and then rotate the pan-tilt camera for tracking people accurately.

This paper is organized as follows. Section II explains, in detail, the algorithm for human detection and also introduces the multi-sensor data fusion. Then, Section III presents several experiments and analyzes the results. Finally, conclusions and future work are illustrated in Section IV.

## II. MOVING OBJECT DETECTION

The moving object detection algorithm adopts multi-sensor data fusion techniques to integrate the following two different sources of information: the first one is body detection, based on the PIR sensors, and the other one is moving object detection, which uses a pan-tilt camera.

*A. Detection based on PIR sensors*

In order to detect the moving object such as human, we propose a global monitoring approach by using 3 PIR sensors that the angle of detection are degree of 120, where PIR's instruction shows the following table and figure.

| Sensor Type | Passive Infrared Range, SK-11 |
|---|---|
| Power | 9 ~ 12V DC |
| Minimum moving object speed | 0.1m/s |
| Maximum moving object speed | 4m/s |
| Possible detection space | 12m, 120° |

Table 1. PIR's specification

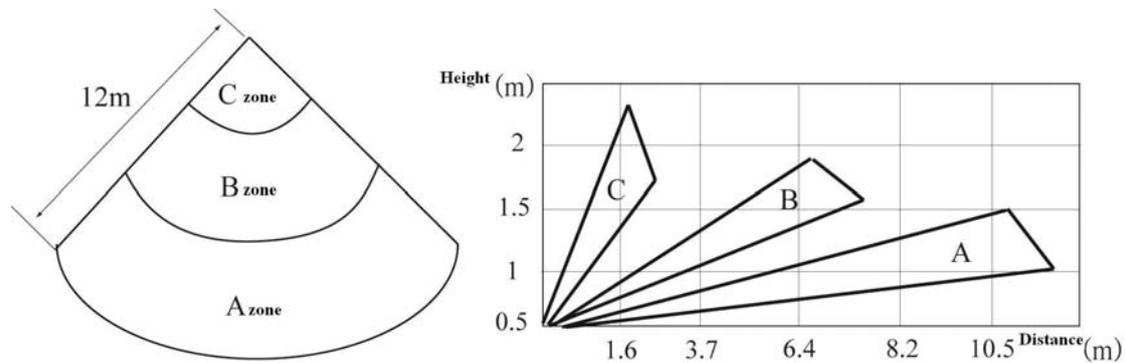

Figure 1. possible detection range

The PIR's detection range can be devided by 3 zones, where A zone is possible of detecion for object with less than 12 *m* in distance and 1.5 *m* high, and C zone is one with less than 2.5 *m* high and 3 *m* in distance .

The moving object detection based on PIR sensor can be expressed by LED color, and if the moving object detect, LED color is red and if no object, LED color is blue. The presence information of moving object transmit the PC computer by A/D converter.

*B. Detection based on frame difference*

The camera motion of the input frame is compensated and consecutive temporal difference is performed to extract moving areas from the image. The moving areas include the target areas (moving objects we are interested in) and some uninteresting motion areas (the moving background).

Since shadows won't have large change between two consecutive frames and little change of shadows to be detected can be removed by the post treatments, shadows have little effect on the accuracy of detection. Thus, they are not handled here to improve the efficiency of this method.

To perform this method in real-time and with high accuracy, we design every block carefully.

Since the consecutive temporal difference approach requires no background model and little memory, this approach is quite efficient and accurate in that it has a low computational cost and it adapts quickly to the changes of the background.

The improved consecutive temporal difference approach is used to quickly obtain moving areas of the input frame. This approach makes use of three consecutive frames. The three frames are divided into two groups. The first group includes the two previous frames (the two consecutive frames that go before the input frame), while the second group includes the input frame and the frame before it. By subtracting the two groups separately, we get two results of different areas from the two subtractions. Since the intersection of the two results is just the very part of the moving area in the frame previous to the input frame, we can obtain the moving areas by subtracting the second results by the intersection of the two difference frames.

The implementation of the method is described in the following.

Let $F_I(k-2), F_I(k-1)$ and $F_I(k)$ represent three consecutive frames, with $F_I(k-2)$ comes earliest and $F_I(k)$ (representing the *kth* input frame) comes latest.

First, we divide the three frames into two groups, with $F_I(k-2)$ and $F_I(k-1)$ a group, $F_I(k-1)$ and $F_I(k)$ another group. Second, we perform subtractions of the two groups separately to get two difference frames.

Suppose a frame has *m* rows and *n* columns, and the value of the pixel at position $(i, j)$ in the *kth* input frame is $f(k,i,j)$, and $F(k)$ can be represented as:

$$F(k) = \begin{pmatrix} f(k,1,1) & \cdots & f(k,1,n) \\ \vdots & & \vdots \\ f(k,m,1) & \cdots & f(k,m,n) \end{pmatrix}, \quad k = 1,2,3,\cdots \qquad (1)$$

Then we obtain two difference frames by subtracting the two groups separately:

$$\begin{cases} D(k) = F(k) - F(k-1) = \begin{pmatrix} |f(k+1,1,1) - f(k,1,1)| & \cdots & |f(k+1,1,n) - f(k,1,n)| \\ \vdots & & \vdots \\ |f(k+1,m,1) - f(k,m,1)| & \cdots & |f(k+1,m,n) - f(k,m,n)| \end{pmatrix} \\ D(k-1) = F(k-1) - F(k-2) = \begin{pmatrix} |f(k,1,1) - f(k-1,1,1)| & \cdots & |f(k,1,n) - f(k-1,1,n)| \\ \vdots & & \vdots \\ |f(k,m,1) - f(k-1,m,1)| & \cdots & |f(k,m,n) - f(k-1,m,n)| \end{pmatrix} \end{cases}$$

$$(2)$$

Then we define the value of every pixel in the difference frames as:

$$\begin{cases} d(k,i,j) = \begin{cases} 0, & |f(k,i,j) - f(k-1,i,j)| \le Th \\ 1, & otherwise \end{cases} \\ d(k-1,i,j) = \begin{cases} 0, & |f(k-1,i,j) - f(k-2,i,j)| \le Th \\ 1, & otherwise \end{cases} \end{cases} \qquad (3)$$

By performing the subtraction above, we get two difference frames, with the pixels with value 0 the background pixels and the pixels with value 1 belonging to moving areas.

Here, $D(k)$ record the moving areas in both $F_I(k)$ and $F_I(k-1)$, while $D(k-1)$ record the moving areas in both $F_I(k-1)$ and $F_I(k-2)$.

Second, we intersect the two diffrence frames obtained in the first step to get the intersection of the pixels with value 1. Since the moving areas in $F_I(k-1)$ are recorded in both of the two difference frames, by intersecting $D(k)$ and $D(k-1)$, we can easily get the moving areas in $F_I(k-1)$.

The moving area in $F_I(k-1)$, that is $M(k-1)$, is calculated like this:
$$M(k-1) = D(k) \cap D(k-1)$$

Third, we obtain the moving areas in $F_I(k)$ by subtracting $D(k)$ from $M(k-1)$.

As we have said before, $D(k)$ contains both the moving areas in $F_I(k)$ and $F_I(k-1)$, and $M(k-1)$ is the moving areas in $F_I(k-1)$, by subtracting $D(k)$ from $M(k-1)$, we get the very moving areas in $F_I(k)$.

$$M(k) = D(k) - M(k-1)$$

*C. Thresolding*

In order to implement the thresolding algorithm on a basis of the concept of similarity between gray levels, we make the following assumptions:

i) there exists a significant contrast between the objects and background:

ii) the gray level is the universe of discourse, a one-dimensional set:

Our purpose is to threshold the gray-level histogram by splitting the image histogram into two crisp subsets, object subset and background subset , using the measure of fuzziness previously defined. Now, based on the assumption i), let us define two linguistic variables {object, background} modeled by two fuzzy subsets, denoted by **B** and **W**, respectively. The fuzzy subsets are associated with the histogram intervals $[x_{min}, x_j]$ and $[x_r, x_{max}]$, respectively, where $x_j$ and $x_r$ are the final and initial gray-level limits for these subsets, and $x_{min}$ and $x_{max}$ are the lowest and highest gray levels of the image, respectively.

We define a *fuzzy region* placed between **B** and **W**, as depicted in Fig. 2. Then, to obtain the segmented version of the gray-level image, we have to classify each gray level of the fuzzy region as being object or background.

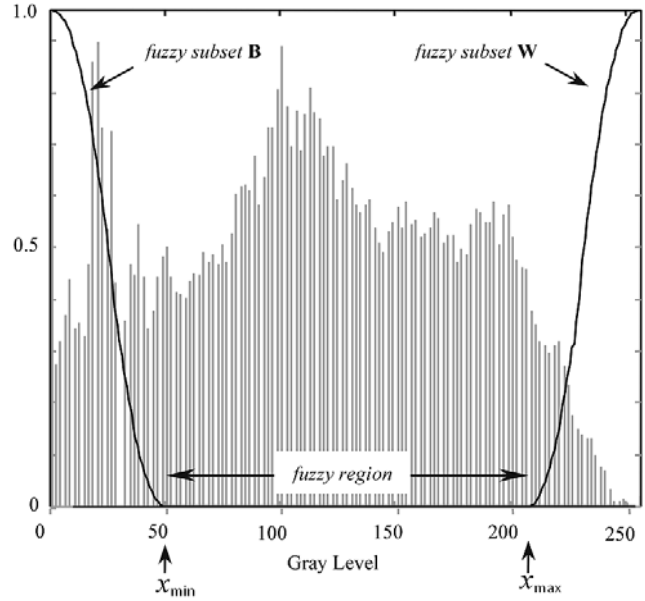

Fig. 2. Multimodal image histogram and the characteristic functions for the seed subsets.

Let us now consider the fuzzy Subsets W And B with membership functions $\mu_B$, $\mu_W$ modeled as follows.

$$\mu_B(x) = \begin{cases} 0, & x < a \\ 2[(x-a)/(c-a)]^2, & a \leq x \leq b \\ 1 - 2[(x-c)/(c-a)]^2, & b < x \leq c \\ 1, & x > c \end{cases} \quad (4)$$

$$\mu_W(x) = 1 - \mu_B(x) \tag{5}$$

Let us take the parameters of the membership functions as follows:

$$b = \frac{\sum_{i=p}^{q} x_i h(x_i)}{\sum_{i=p}^{q} h(x_i)}, \quad a = 2b - c \tag{6}$$

$$c = b + \max\{|b - x_{\max}|, |b - x_{\min}|\}$$

, where $h(x_i)$ denotes the image histogram and $x_p$ and $x_q$ are the limits of the subset being considered.

The proposed algorithm to get the thresold can be summarized in the following Steps:

[*Thresolding Algorithm*]

Step 1: For histogram level $i$ ($x_{\min} \leq i \leq x_{\max}$), the distance calculate as follows:

$$\mu_{\bar{A}}(x_i) = \begin{cases} 0, & \text{if } \mu_A(x_i) < 0.5 \\ 1, & \text{if } \mu_A(x_i) \geq 0.5 \end{cases} \tag{7}$$

$$d_k(A, \bar{A}) = \frac{2}{n^{1/k}} \left( \sum_{i=1}^{n} (\mu_A(x_i) - \mu_{\bar{A}}(x_i))^k \right)^{1/k} \tag{8}$$

$$\psi_k(A) = \frac{2}{n^{1/k}} d_k(A, \bar{A}) \tag{9}$$

Step 2: compute the normalization factor $\alpha$.

$$\alpha = \frac{\psi_k(\mathbf{W})}{\psi_k(\mathbf{B})} \tag{10}$$

Step 3: The thresold level for image segmentation is determined by the intersection of the normalized curves of the indices of fuzziness.

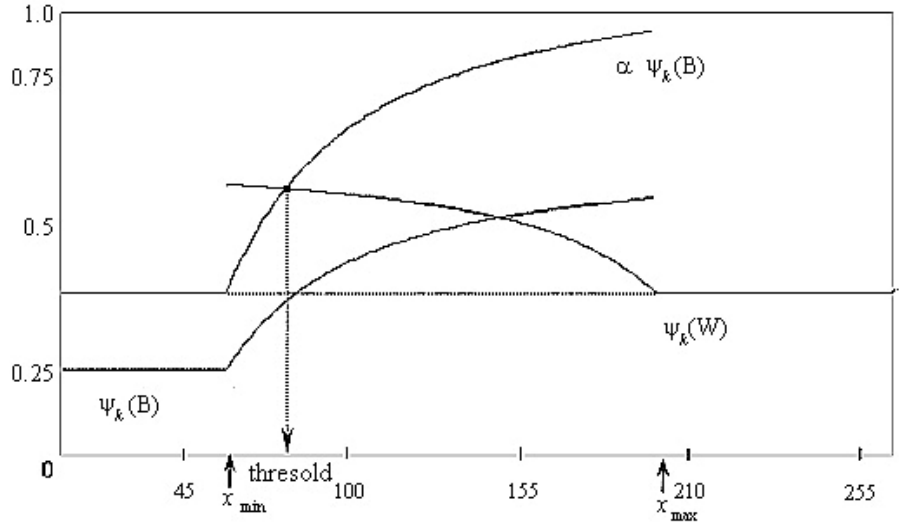

Fig. 3. Fuzziness and determination of the threshold value.

*D. Sensor fusion*

The mobile robot with multi-sensors arise the important problem how to combine the sensor's information. As the PIR's angle of detection is large than the pan-tilt camera's one, we propose the rules combining sensors as follows.

[*Rule for Sensor fusion*]

Rule 1: if "*Camera found*" Then "*Camera Tracking*"

Rule 2: if "*Infer1 found*"   Then
        If  $\alpha \leq -45$   Then "*Camera Turn to Rght*"
        If  $\alpha \geq 45$    Then "*Camera Turn to Left*"
        Else "*Camera Tracking*"

Rule 3: if "*Infer2 found*" Then
        If  $\alpha \leq -120$   Then "*Camera Turn to Rght*"
        If  $\alpha \geq -60$    Then "*Camera Turn to Left*"
        Else "*Camera Tracking*"

Rule 4: if "*Infer3 found*" Then
        If  $\alpha \leq 60$    Then "*Camera Turn to Rght*"
        If  $\alpha \geq 120$   Then "*Camera Turn to Left*"
        Else "*Camera Tracking*"

Rule 5: if "*Infer1 found*" and "*Infer2 found*" Then
        If  $\alpha < -90$    Then "*Camera Turn to Rght*"
        If   $\alpha > 0$        Then "*Camera Turn to Left*"
        Else "*Camera Tracking*"

Rule 6: if "*Infer1 found*" and "*Infer3 found*" Then
        If $\alpha < 0$     Then "*Camera Turn to Rght*"
        If  $\alpha > 120$  Then "*Camera Turn to Left*"

Else "*Camera Tracking*"

Rule 7: if "*Camera not found*" and "*Infer*1 not *found*" and "*Infer*2 not *found*" and "*Infer*3 not *found*" Then "*Camera Turn to Zero position*"

Above rules, *Infer1,Infer2,Infer3* means the PIR sensors, and $\alpha$ is a pan-tilt camera's angle of rotation. "*Camera Turn to Rght*" or "*Camera Turn to Left*" means that a pan-tilt camera turns to right or left with a defined speed of stepping motor, and "*Camera Turn to Zero position*" means that the camera turns to the initial position so that the camera captures object in front of the mobile robot. Also, "*Camera Tracking*" means that the camera turns so that the center of target(human detected) coincedent with the center of image coordinates.

## Ⅲ. EXPERIMENTAL RESULTS

To test the performance and the portability of the proposed solution, the system has been implemented on mobile robot "RYONG NAM SAN-No 1" shown in Fig. 1, which is provided with a PIR sensor and a Pan-Tilt camera. This is mounted on a special support at approximately 0.8 m from the floor in order to facilitate the face detection. The onboard computer is a Core Duo 1.66 GHz with 1 GB of RAM. A touch screen is also available for interaction.

The proposed mobile security robot runs on a windows operating system, and the whole software has been written in Microsoft VC++ and runs in real time on the robot PCs, although it is possible to use an external client, connected via wireless, for remote control and debug. The camera provide images with a resolution of 320 $\times$ 240 pixels at 33 Hz.

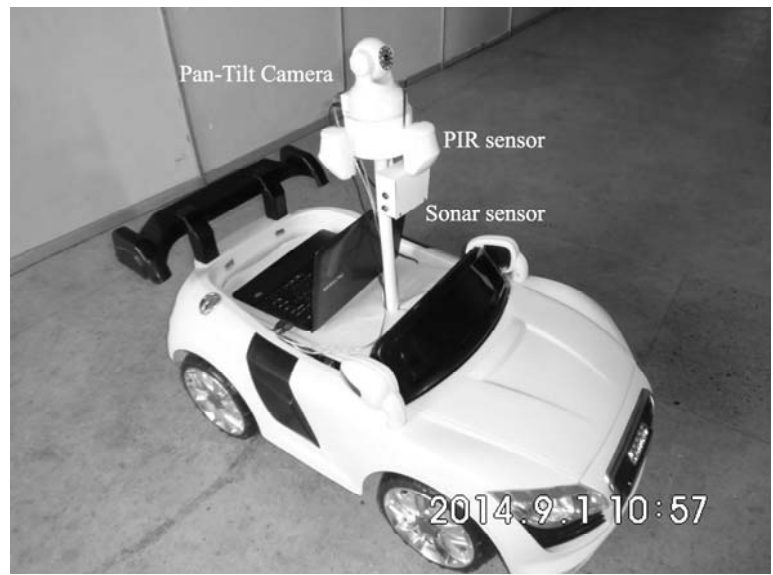

Fig 4. RYONG NAM SAN-No1

The proposed mobile robot consists of mobile robot, 3 infrared range sensors, a pan-tilt camera, and laptop as the following figure.

The walking speed constraints of human detection have been empirically determined after analyzing many recorded data of different people walking in typical indoor environments. The best results have been obtained setting the minimum speed to 0.1m/s, the maximum speed to 4m/s, and the width of body is 48cm. The experiments have been conducted in our laboratory, a robot arena, and adjacent offices, as shown in Fig. 5, moving between different rooms and, often, crossing doorways or narrow corridors.

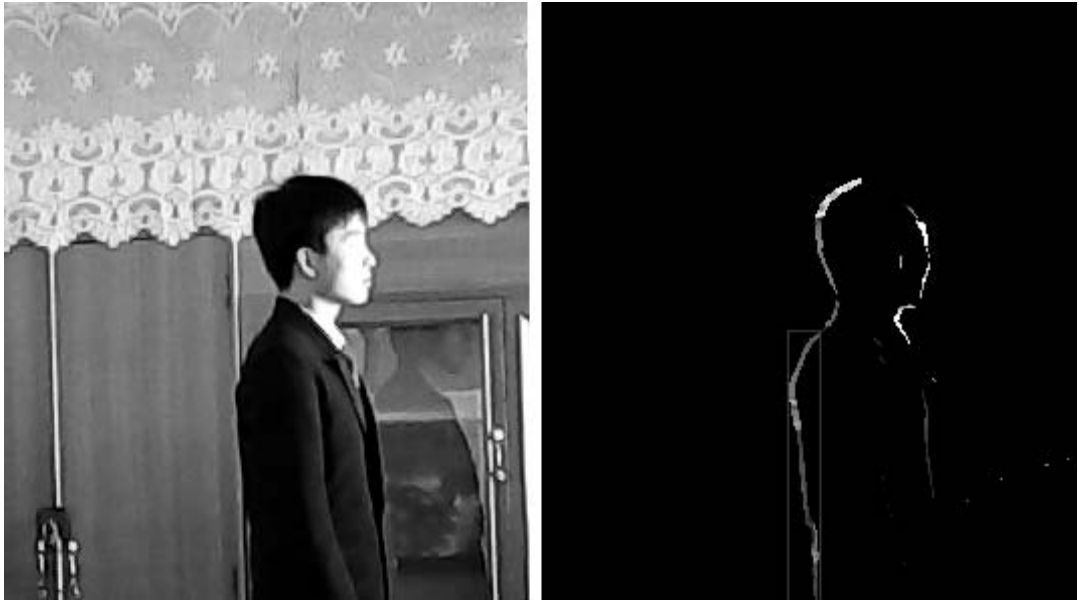

Fig 5. Experimental results of moving objects detection

During the experiments, the robots were controlled remotely to follow the persons being tracked, often moving faster than 0.5 m/s and with a turn rate of up to 45 °/s.

## IV. CONCLUSIONS

Based on consecutive temporal difference and the fuzzy clustering algorithm, we propose a method for motion detection in complex background in traffic monitoring systems, which is proved to be efficient, accurate and robust. Compared with other similar motion detection algorithms, the main improvement of the proposed method is that it requires only a little time and memory, thus suitable for use in real-time applications. In addition, no prior knowledge of the background is needed for the implementation of this method, and it is quite robust to background changes, not accumulating previous mistakes. Tests on the standard data sets also demonstrate that it has an outstanding performance.

## References

[1] Nicola Bellotto, Huosheng Hu "Multisensor-Based Human detection and Tracking for Mobile Service Robots", IEEE Trans on Systems,Man, And Cybernetics, Vol.39, No.1, 2009, pp.167-182

[2] W. Burgard, P. Trahanias, D. Hähnel, M. Moors, D. Schulz, H. Baltzakis, and A. Argyros, "TOURBOT and WebFAIR: Web-operated mobile robots for tele-presence in populated exhibitions," in *Proc. IROS Workshop Robots Exhib.*, 2002, pp. 1–10.

[3] J. N. K. Liu, M. Wang, and B. Feng, " iBotGuard: An Internet-based intelligent robot security system using invariant face recognition against intruder," *IEEE Trans. Syst., Man, Cybern. C, Appl. Rev.*, vol. 35, no. 1, pp. 97–105, Feb. 2005.

[4] Durus M., Ercil A., "Robust Vehicle Detection Algorithm," Signal Processing and Communications Applications, 2007. SIU 2007. IEEE 15th 11-13 June 2007 pp.1 – 4

[5] Fu-Yuan Hu, Yan-Ning Zhang, Lan Yao, "An effective detection algorithm for moving object ith complex background," Machine Learning and Cybernetics, 2005. Proceedings of 2005 International Conference on Volume 8, 18-21 Aug. 2005 pp:5011 - 5015 Vol. 8

[6] Yeon-sung Choi, Piao Zaijun, Sun-woo Kim, Tae-hun Kim, Chun-bae Park, "Motion Information Detection Technique Using Weighted Subtraction Image and Motion Vector," Hybrid Information Technology, 2006. ICHIT '06. Vol 1. International Conference on Volume 1, Nov. 2006 pp:263 – 269

[7] Ho M.A.T., Yamada Y., Umetani Y., "An HMM-based temporal difference learning with model-updating capability for visual tracking of human communicational behaviors," Automatic Face and Gesture Recognition, 2002. Proceedings. Fifth IEEE International Conference on 20-21 May 2002 pp:163 –

168

[8] Lim S., Apostolopoulos J.G., Gamal A.E., "Optical flow estimation using temporally oversampled video," Image Processing, IEEE Transactions on Volume 14, Issue 8, Aug. 2005 pp:1074 – 1087

[9] Qi Zang , "a Moving Object Using One-Dimensional Optical Flow with a Rotating Observer," Control, Automation, Robotics and Vision, 2006. ICARCV '06. 9th International Conference on 5-8 Dec. 2006, pp:1 – 6

[10] Zhen Tang, Zhenjiang Miao, "Fast Background Subtraction and Shadow Elimination Using Improved Gaussian Mixture Model," Haptic, Audio and Visual Environments and Games, 2007. HAVE 2007. IEEE International Workshop on 12-14 Oct. 2007 pp:38 –41

[11] Qi Zang, Klette R., "Robust background subtraction and maintenance," Pattern Recognition, 2004. ICPR 2004. Proceedings of the 17th International Conference on Volume 2, 23-26 Aug. 2004 pp:90 - 93 Vol.2

[12] M. Lindström and J.-O. Eklundh, "Detecting and tracking moving objects from a mobile platform using a laser range scanner," in *Proc. IEEE/RSJ Int. Conf. Intell. Robots Syst.*, Maui, HI, 2001, vol. 3, pp. 1364–1369.

[13] W. Zajdel, Z. Zivkovic, and B. J. A. Krose, "Keeping track of humans: Have I seen this person before?" in *Proc. IEEE Int. Conf. Robot. Autom.*, Barcelona, Spain, 2005, pp. 2093–2098.

[14] Zhen Yu, and Yanping Chen, " A real-time motion detection algorithm for traffic monitoring systems based on consecutive temporal difference", Proceedings of the 7[th] Asian Control Conference, Hong Kong, China, August 27-29, 2009, pp. 1594-1600

[15] Masahiro Yamada, Chi-Hsien Lin and Ming-Yang Cheng, "Vision Based Obstacle Avoidance and Target Tracking for Autonomous Mobile Robots", The 11[th] IEEE International Workshop on Advanced Motion Control March 21-24, 2010, Nagao, Japan, pp 153-159

[16] Thomas Nierobisch, Wladimir Fischer and Frank Hoffmann, " Large View Visual Servoing of a Mobile Robot with a Pan-Tilt Camera", Proceedings of the 2006 IEEE/RSJ International Conference on Intelligent Robots and Systems October 9-15,2006,pp. 3307-3313

[17] M.A.El-Bardini, E.A.Elsheikh and M.A.Fkirin, " Real Time Object Tracking Using Image Based Visual Servo Technique", International Journal of Computer Science & Emerging Technologies Volume 2, 2011, pp.252-258

[18] Himanshu Borse, Amol Dumbare, and Rohit Gaikwad, " Mobile Robots for Object Detection Using Image Processing", Global Journal of Computer Science and Technology Neural & Artificial Intelligence Volume 12, 2012, pp. 13-17

[19] Meng Shiun Yu, Horng Wu, and Huei Yung Lin, " A Visual Surveillance System for Mobile Robot using Omnidirectional and PTZ cameras", SICE Annual Conference 2010, pp. 37-43

[20] Sayana Sivanand, "Adaptive Local Thresold Algorithm and Kernel Fuzzy C-Means Clustering Method for Image Segmentation", International Journal of Latest Trends in Engineering and Technology Vol 2, 2013, pp. 261-266